\documentclass{article}
\usepackage[preprint]{neurips_2021}

\usepackage[utf8]{inputenc} % allow utf-8 input
\usepackage[T1]{fontenc}    % use 8-bit T1 fonts
\usepackage{hyperref}       % hyperlinks
\usepackage{url}            % simple URL typesetting
\usepackage{booktabs}       % professional-quality tables
\usepackage{amsfonts}       % blackboard math symbols
\usepackage{nicefrac}       % compact symbols for 1/2, etc.
\usepackage{microtype}      % microtypography
\usepackage{xcolor}         % colors
\usepackage{multicol}
\usepackage{multirow}
\usepackage{wrapfig}

\usepackage{amsmath,amsthm,mathtools,amsfonts,amssymb}
\usepackage{graphicx}
\usepackage{algorithm, algorithmic}
\usepackage{subfigure,subfloat}
\usepackage{soul}

\title{Efficient physics-informed neural networks using \\hash encoding}

\author{%
Xinquan Huang$^{1}$, Tariq Alkhalifah$^{1}$\\
$^1$King Abdullah University of Science and Technology\\
\texttt{\{xinquan.huang, tariq.alkhalifah\}@kaust.edu.sa}\\
}

\begin{document}
\maketitle

\begin{abstract}
Physics-informed neural networks (PINNs) have attracted a lot of attention in scientific computing as their functional representation of partial differential equation (PDE) solutions offers flexibility and accuracy features. However, their training cost has limited their practical use as a real alternative to classic numerical methods. Thus, we propose to incorporate multi-resolution hash encoding into PINNs to improve the training efficiency, as such encoding offers a locally-aware (at multi resolution) coordinate inputs to the neural network. Borrowed from the neural representation field community (NeRF), we investigate the robustness of calculating the derivatives of such hash encoded neural networks with respect to the input coordinates, which is often needed by the PINN loss terms. We propose to replace the automatic differentiation with finite-difference calculations of the derivatives to address the discontinuous nature of such derivatives. We also share the appropriate ranges for the hash encoding hyperparameters to obtain robust derivatives. We test the proposed method on three problems, including Burgers equation, Helmholtz equation, and Navier-Stokes equation. The proposed method admits about a 10-fold improvement in efficiency over the vanilla PINN implementation.
\end{abstract}

\section{Introduction}
\label{introduction}
Partial differential equations (PDEs) are essential in science and engineering as they represent physical laws that describe basic natural phenomena, like heat transfer, fluid flow, and wave propagation, with applications in optimal control, medical and Earth imaging, and inversion. 
However, conventional methods for solving PDEs, e.g., finite-difference, finite-element, or spectral-element methods, often require complex discretization, and intensive computation, and are prone to numerical errors. 
These limitations are unfavorable to inverse design and implementation on regions of complex geometry. 
With the recent developments in computational resources and the availability of robust machine learning frameworks, scientific machine learning has taken center stage,
especially in tasks related to solving PDEs. These solutions are often learned in a supervised manner
using numerically generated labels for the training \citep{guo_convolutional_2016,zhu_bayesian_2018}. However, recently,
physics-informed neural networks (PINNs) \citep{Raissi2019} and operator 
learning~\citep{lu_deeponet_2021,li_fourier_2020} have shown their potential to revolutionize scientific computing. While operator learning focuses on representing the kernel that transforms inputs to outputs, like learning a PDE solver for many PDE parameters (included in the training), PINN is meant to learn the functional solution of a specific PDE and is not inherently designed to be applied to various PDE parameters, unless transfer or meta-learning is involved and that is possible when the changes in PDE parameters are small \citep{goswami_transfer_2020,qin_meta-pde_2022}. Since PINN training takes up the role of inference in machine learning, the efficiency of the training is crucial, which is not the case. Despite this fundamental limitation, the quest to solve this problem has only intensified, as the functional representation of the PDE solutions offers all kinds of flexibility and accuracy features.

In PINNs, we are constructing neural network functional representations of the solutions of PDEs that would otherwise be defined on a fixed mesh and computed through numerical methods, like finite difference or finite element. The functional representation (an approximation) offers a solution in a form that would only be available if we could solve the PDE analytically. The neural network, $f({\bf x})$, where ${\bf x}$ represents the coordinates of the domain of interest offers flexibility in representation in irregular domains, and domains with gaps, and offers accuracy beyond the numerical approximations, especially in computing the derivatives of the function using automatic differentiation \citep{baydin_automatic_2018}. This functional representation of the solution of a PDE is attained through a neural network optimization problem that could be costly. Conventional PINN training consumes 10000s of epochs at a cost that is often much higher than traditional numerical methods, which makes the use of PINNs for practical applications, beyond solving exotic equations, less attractive. The primary reason behind the high cost of PINN is the number of epochs needed to converge, especially for complex functional solutions \citep{wang_eigenvector_2021}. Some of these limitations can be attributed to the low-frequency bias of neural networks \citep{Rahaman2018,xu_frequency_2019,huang_pinnup_2022}. It takes the multi-layer perceptron, initialized randomly, a while of training before it can find a path toward a potential local minimum (sometimes thousands of epochs). In other words, it is first lost in a random search, considering the often high dimensional nature of the neural network space, before it finds its footing \citep{qin_meta-pde_2022}. Part of the problem is the small imprint that the solution coordinates values (inputs), limited by its dimension (often three space and time), can exert on the neural network, which forces that network to initially roam freely in the parameter space with little guidance. This limitation has been mildly addressed with increasing the influence of inputs through encoding, which allows for higher dimensional representation of the input space, in which the scalar inputs are replaced by vectors. An example, given by frequency or positional encoding, replaces the small difference in coordinates for neighboring samples to more profound changes in the network represented by bigger differences in the positional vector representation \citep{liu_neural_2021,takikawa_neural_2021,Huang2021}.

To improve the efficiency of PINNs training, three components of the PINNs machinery have been addressed: the neural network (NN) architecture design, the loss function design, and the training process \citep{cuomo_scientific_2022}. In this paper, we focus on the NNs design, and specifically in representing the input (with an embedding layer). To some extent, PINNs could be regarded as a neural field representation task (e.g., NeRF), whose inputs are spatial coordinates and outputs are the voxels or the physical fields, but with a PDE loss as a training signal. Inspired by the success of embedding methods, and specifically, the encoding of the input coordinates to the neural field representation, and the recent huge progress on NeRF given by multi-resolution hash encoding \citep{muller_instant_2022}, prompted a logical question: {\it can we leverage hash encoding to accelerate the PINNs training?} So, hash encoding, used as a form of encryption over the years, of the input coordinates can provide a more locally-aware representation of the coordinate values. This feature over multi-resolution provided considerable acceleration in the convergence of neural network functional representations of images (NeRF). Convergence rates of 1000s of epochs were reduced to below 100. This is a major improvement that allowed for the practical use of NeRFs. However, the supervised training involved in such representations did not require the calculation of derivatives as is needed by the PDE loss function in PINNs.

In this paper, we introduce hash encoding to PINNs. We investigate its capability in solving the outstanding cost limitation of PINN by reducing the number of epochs needed to converge. We also investigate the applicability of automatic differentiation through hash encoding. Alternatively, we use the finite-difference method to ensure the stable calculation of the derivatives of the NN function with hash encoding and use it in the PINNs training. In the numerical examples, we test our approach on several PDEs to demonstrate its potential in considerably reducing the cost of PINNs, as well as share the limitations we encountered.

The main contributions of this study are the following:
\begin{itemize}
    \item We propose an efficient physics-informed neural network by means of hash encoding.
    \item We make use of the finite difference method to obtain the first and second-order derivatives and avoid the influence of discontinuous derivatives on automatic differentiation.
    \item We validate our method on the three PDE boundary value problems, including Burgers equation, Helmholtz equation, and Navier-Stokes equation, and achieve 10-fold acceleration in PINNs training.
\end{itemize}

In the following sections, we first briefly summarize the related works in Section~\ref{related_work} and then introduce the preliminaries and the proposed PINN using hash encoding in Section~\ref{methodology}. To showcase the efficiency and accuracy of the proposed method, we present the settings of the experiment and results in Section~\ref{experiments}. Finally, we conclude and discuss potential future work in Section~\ref{conclusion}.

\section{Related Work} 
\label{related_work}
\subsection{Physics-informed neural networks}
The concept of using a neural network functional representation to solve PDEs was first introduced in the 20th century and was validated by the universal approximation theorem \citep{Hornik1989,Lagaris1998}. 
\cite{Raissi2019} introduced the physics-informed neural network (PINN) framework and showed its application in fluids, quantum mechanics, reaction–diffusion systems, and the propagation of nonlinear shallow-water waves. 
The general idea of PINNs is to train a mapping function from the input coordinates (spatial coordinates and/or time) to the output physical field, which satisfies the physical governing equation. The loss function is the PDE residuals and any initial or boundary conditions; thus, it is regarded as an unsupervised technique.
\cite{Alkhalifah2020,Sitzmann2020,huang_pinnup_2022} showed its potential as an efficient surrogate modeling approach for frequency-domain wavefields. 
PINN solutions can adapt to any model shape, including irregular topography and interior geometry. However, it is trained to provide the solution for specific PDE parameters, and thus, it requires retraining or transfer learning if the PDE parameters change \citep{goswami_transfer_2020}, which limits its rapid use as a numerical solver of PDEs and varying parameters, like those we encounter in an inversion process. The model architecture, the training samples, the loss function, and even the initialization of the NN all have
distinct affect on the convergence of PINNs. 
\cite{wu_comprehensive_2023} proposed residual-based adaptive distribution and residual-based adaptive refinement with distribution to improve the sample efficiency during training. 
\cite{huang_single_2022} proposed a single reference frequency loss function and \cite{huang_pinnup_2022} proposed frequency upscaling and neuron splitting to help the PINN solve the Helmholtz equation converge
at high frequencies. 
\cite{sharma_accelerated_2022} proposed meshless discretizations for derivatives calculation to accelerate the training of PINNs.
\cite{qin_meta-pde_2022} proposed Meta-PDE, which involves using gradient-based meta-learning to amortize the training time needed to fit the NN on a problem drawn from a distribution of parameterized PDEs, and achieves an intermediate accuracy approximation in up to an order of magnitude speedup. However, even with those promising developments, there is still a long way to go to the ultimate goal to surrogate the numerical simulation with neural networks.

\subsection{Input Encoding}
The objective of PINNs is to train an NN function of coordinate inputs to output a solution that respects the physical laws. 
It has been shown that the success of such a task relies heavily on the embedding that maps the input of the NNs to a higher-dimensional space (positional encoding). 
Early examples of encoding the input of an NN, training-free encoding, include the basic one-hot encoding \citep{harris_combinational_2013}, the kernel trick \citep{theodoridis_pattern_2006}, and later, the implementation of positional encoding using sine and cosine functions \citep{vaswani_attention_2017}. The latter approach has resulted in convergence improvements in PINNs \citep{Huang2021}. 
\cite{muller_neural_2020} developed the one-blob encoding, a continuous variant of the one-hot encoding, which shows better performance compared to the encoding using a sinusoidal function.
Compared to these analytic encoding methods, recent progress on parametric encodings, which make use of additional trainable parameters in an auxiliary data structure, like grid or tree, has shown state-of-the-art performances, e.g., grid-based or tree-based encoding \citep{jiang_local_2020,mehta_modulated_2021,martel_acorn_2021,sun_direct_2022,muller_instant_2022}. Among these methods, the multi-resolution hash encoding \citep{muller_instant_2022} has reduced the training cost of NeRF from days to seconds. It reduces the memory access operation and the number of floating point operations by means of a multi-resolution hash table of trainable feature vectors, where values are optimized through stochastic gradient descent, achieving a considerable increase in efficiency. 
Although the utilization of hash encoding has taken computer vision with neural networks to a new era, its potential benefits to PINNs are still unclear and needs to be explored because unlike NeRF, which relies on supervised learning, PINNs are driven by the corresponding PDE requiring derivative calculations of the solution
with respect to the input.
To the best of our knowledge, we are the first to combine hash encoding with physics-informed neural networks with the fundamental purpose of reducing the cost of training PINNs.

\section{Methodology} 
\label{methodology}
In this section, we first review the framework of physics-informed neural networks (PINNs) and the concept behind hash encoding.
Then we investigate the incorporation of hash encoding into PINNs, and specifically analyze options for differentiation including finite difference and automatic differentiation considering the discontinuous nature of the vanilla multi-resolution hash encoding derivatives.
\subsection{Preliminaries}
Considering a connected domain of $n$ dimensions $\Omega \subseteq \mathbb{R}^n$ and boundary $\partial \Omega$, a general time-dependent PDE can be defined as:
\begin{equation}
u_t(\mathbf{x})+S(u(\mathbf{x}), a(\mathbf{x}))=0, \quad \mathbf{x} \in \Omega, \quad t \in[0, T],
\end{equation}
where $t$ and $\mathbf{x}$ are the time and spatial coordinates, respectively, $S(u, a)$ is a non-linear differential operator, and $a\in \mathcal{A}$ represents the parameters of the PDE, e.g., coefficients and initial or boundary conditions, and $u$ represents the physical field we want to solve for. 
In vanilla PINNs, a neural network $\Phi(\theta,\mathbf{x},t)$, parameterized by the trainable parameters $\theta$, is trained to map the input coordinates (including time for time-dependent equations) to the output, which represents the physical field (e.g., velocity, pressure, vorticity, and so on) at the input coordinate location, satisfying the following equation:
\begin{equation}
    \frac{\partial \Phi(\theta,\mathbf{x})}{\partial t} + S(\Phi(\theta,\mathbf{x}), a(\mathbf{x})) = 0.
\end{equation}
Thus, we can use the mean square residual of this PDE, as well as any initial or boundary conditions, in the loss function,
\begin{equation}
    \mathbf{L} = 
\underbrace{\frac{1}{N_i} \sum_{j=1}^{N_i}\left(\frac{\partial \Phi(\theta,\mathbf{x}_j)}{\partial t} + S(\Phi(\theta,\mathbf{x}_j), a(\mathbf{x}_j))
\right)^2}_{\text {Interior PDE loss}}+\underbrace{\frac{1}{N_b} \sum_{i=1}^{N_b}\left(\mathcal{B}(\Phi(\theta,\mathbf{x}_i))-u_{b}\left(\mathbf{x}_i\right)\right)^2}_{\text {Superivsed loss on boundary}},
\end{equation}
to optimize the parameters of the NN, $\theta$, where $N_i$ is the number of collection points in the domain and $N_b$ is that on the boundary, $u_b$ is the boundary values, and $\mathcal{B}$ is the boundary operator, denoting derivatives or values of the field.
\subsection{Hash encoding}
\begin{figure}[!htb]
    \centering
    \includegraphics[width=\columnwidth]{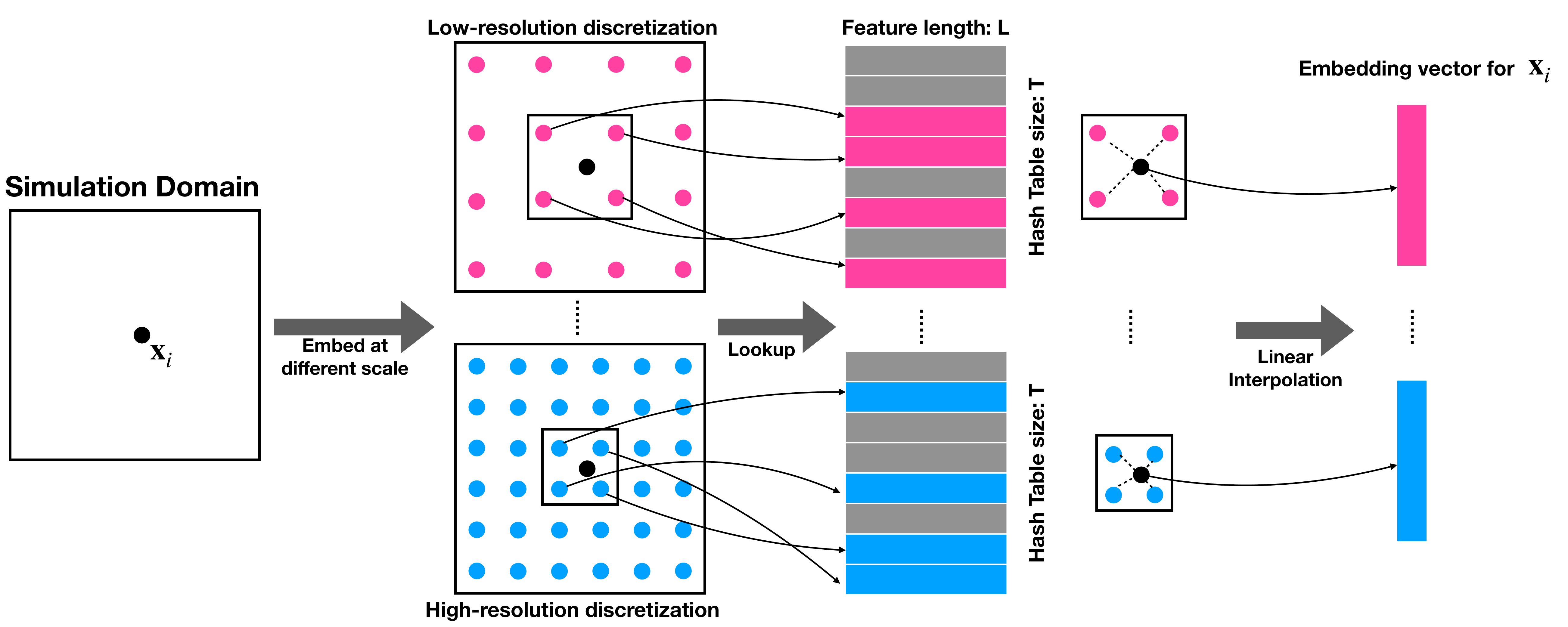}
    \caption{The diagram of the hash encoding, where different colors denote the different scales (resolution) and corresponding embedding vectors. }
    \label{fig:hash}
\end{figure}

For function learning, we aim to improve the approximation quality of the NN to the PDE, and also the speed of training for a given NN. 
Note that speed is the main objective of this paper. 
A smart way is to encode the input query, e.g., the spatial coordinates and time, to a high-dimensional space. 
Here, we use the hash encoding proposed by \cite{muller_instant_2022}. 
The general idea of hash encoding is to combine a multi-resolution decomposition with a hash table to represent 3D shapes compactly. 
Then the complex 3D real world could be represented by a small neural network and trainable hash encoding parameters.
Figure~\ref{fig:hash} includes a diagram depicting the hash encoding mechanism modeled after the diagram used by \cite{muller_instant_2022}.
Each sample in the simulation domain $\mathbf{x}_i$ can be described using $S$ levels of resolution from low to high.
For each level of resolution, like the pink or the blue dots in Figure~\ref{fig:hash}, we calculate the embedding vector. 
Specifically, we first find its voxel vertices (4 vertices for the 2D case, and 8 vertices for the 3D case) and then use a trainable hash table, which includes fixed features of length $L$ and hash table size $T$ for each level of resolution, to evaluate the corresponding embedding vector for each vertex.
We, then, use linear interpolation of the vertices vectors to obtain the embedding vector for $\mathbf{x}_i$ at different levels. 
Finally, the hash encoding for $\mathbf{x}_i$ is the concatenation of these embedding vectors from different levels.

Specifically, given the encoding level $S$, and the finest and coarsest resolution $R_f$ and $R_c$, the resolution of each level $R_s$ is determined by means of geometric progression, as follows:
\begin{equation}
\begin{aligned}
R_s & :=\left\lfloor R_c \cdot b^s\right\rfloor, \\
b & :=\exp \left(\frac{\ln R_f-\ln R_c}{S-1}\right).
\end{aligned}
    \label{equ:resolution}
\end{equation}
Then for $\mathbf{x}_i$, its vertices are $\left\lfloor\mathbf{x}_{i,s}\right\rfloor:=\left\lfloor\mathbf{x}_{i} \cdot R_s\right\rfloor$ and $\left\lceil\mathbf{x}_{i,s}\right\rceil:=\left\lceil\mathbf{x}_{i} \cdot R_s\right\rceil$. 
As for the coarse resolution, where the number of vertices $(R_s+1)^{d}$ is smaller than the hash table $T$, the hash table can provide a one-to-one query.
However, for higher resolution, the mapping between the vertices and the hash table is achieved by a hash function 
\begin{equation}    h(\mathbf{x}_i)=\left(\bigoplus_{j=1}^d x_{i,j} \pi_j\right) \quad \bmod T,
    \label{equ:hash}
\end{equation}
where $\bigoplus$ is a bitwise "exclusive or" XOR, and $\pi_j$ are unique and large prime numbers \cite{teschner_optimized_2003}.
This kind of encoding not only provides a compact representation of the input dimensions but also is quite efficient with a computational complexity of $O(1)$ due to the practically instant hash table lookup mechanism. 

\subsection{PINNs with hash encoding}
\begin{figure}[!htb]
  \centering
  \subfigure[T=10]{\label{hash_ad_10}\includegraphics[width=0.32\textwidth]{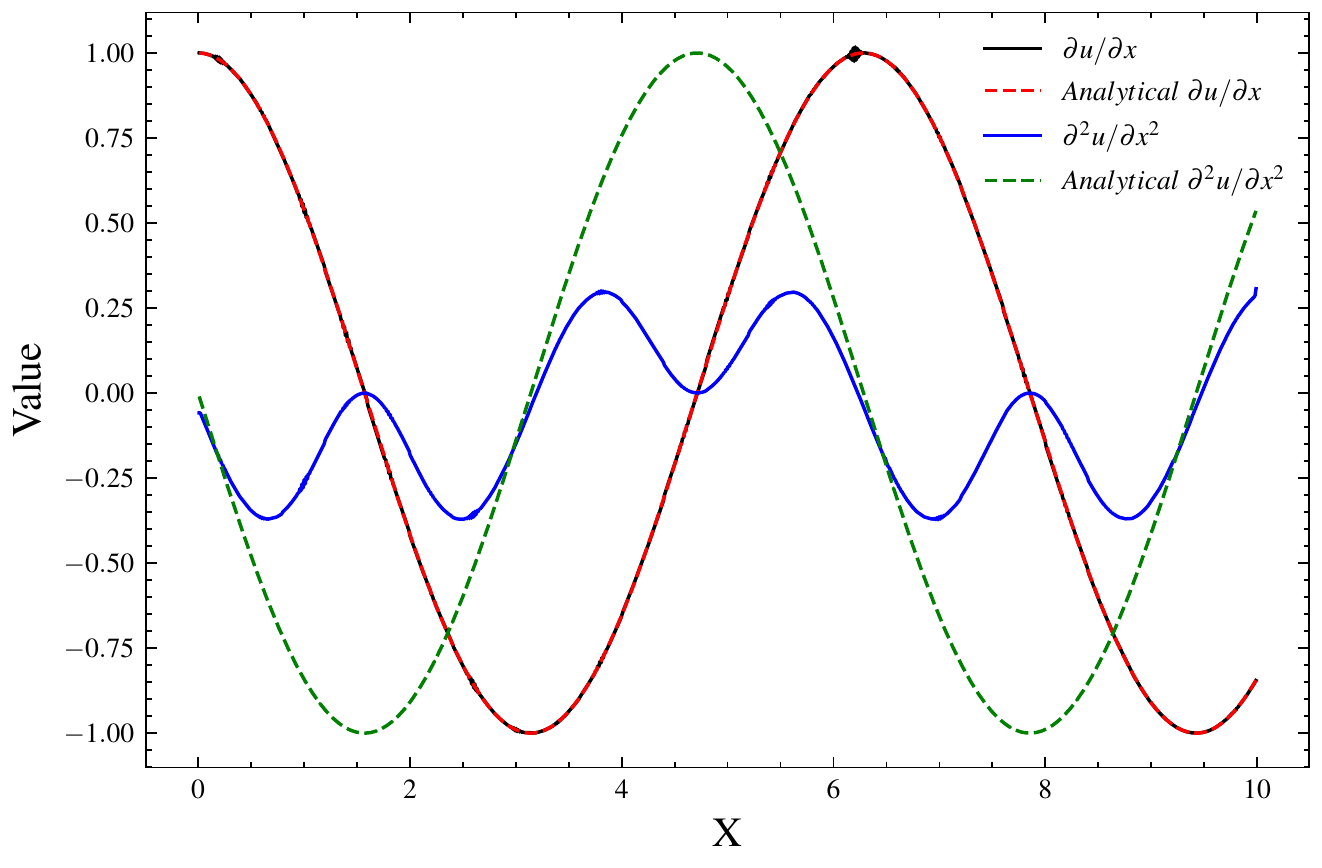}}
  \subfigure[T=8]{\label{hash_ad_8}\includegraphics[width=0.32\textwidth]{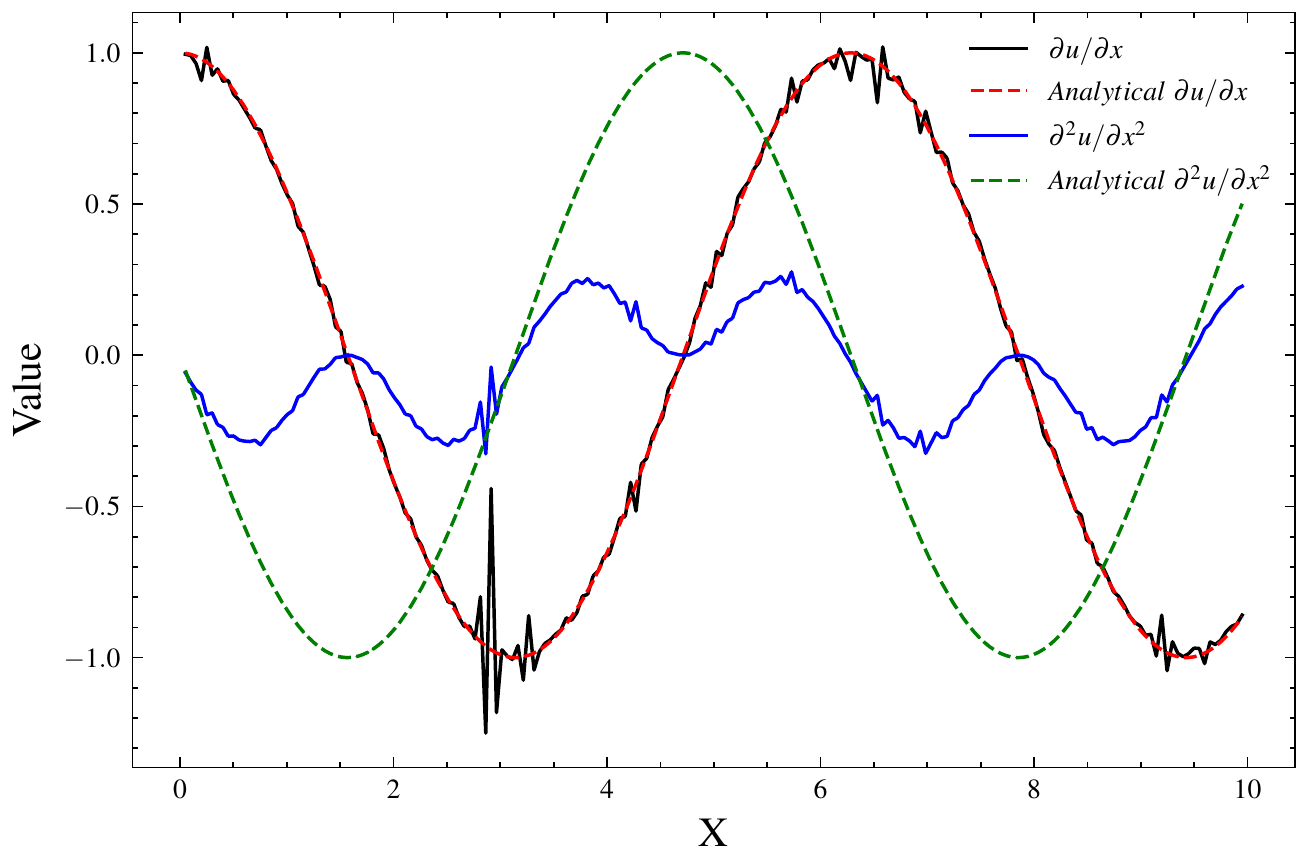}}
    \subfigure[T=4]{\label{hash_ad_4}\includegraphics[width=0.32\textwidth]{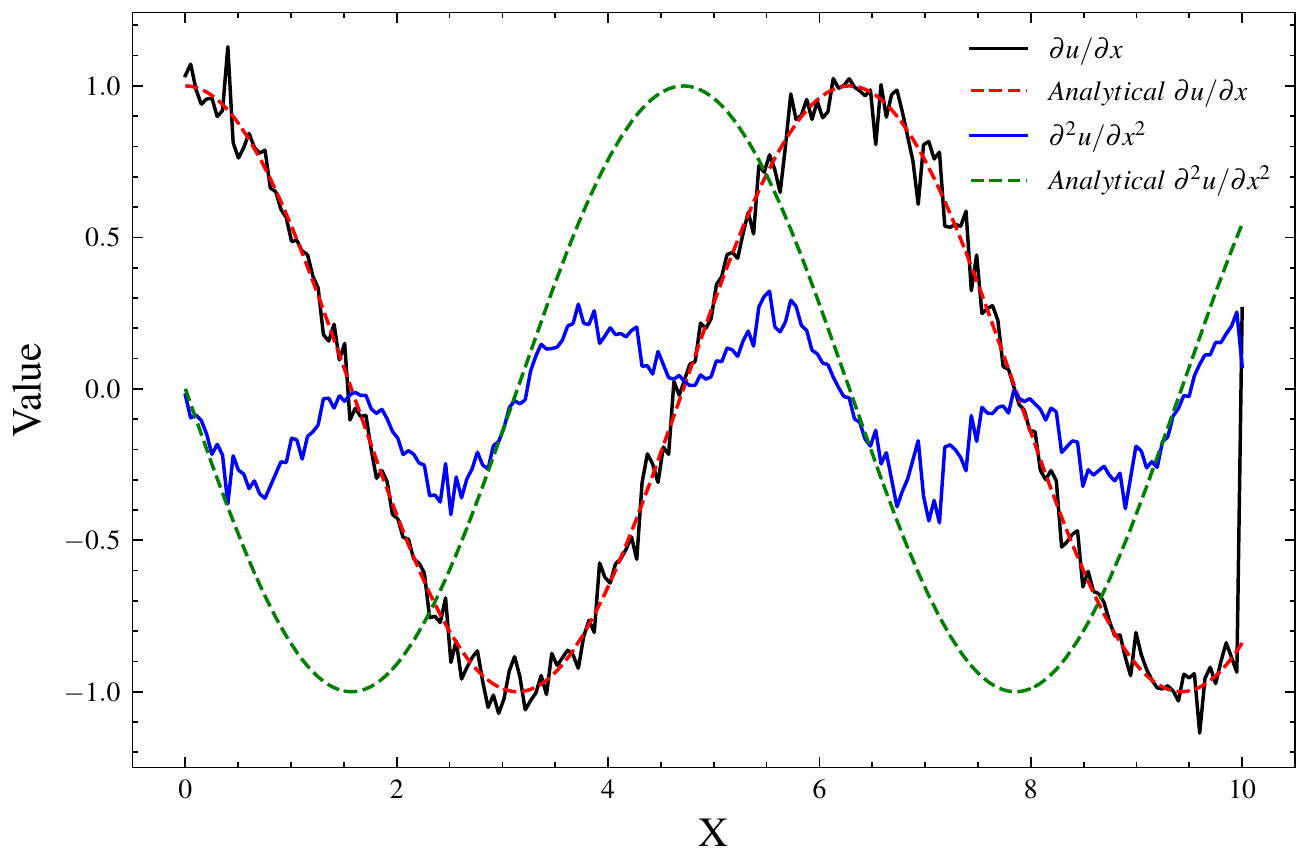}}
    \caption{Illustration of the accuracy of the first- and second-order derivatives calculation by the AD method. We use this NN to fit $f=sin(x)$ with a multi-resolution hash encoding and visualize its first- and second-order derivatives for a hash table size of 10 in (a), and also visualize the derivatives with hash table sizes of 8 and 4 in (b) and (c), respectively.}
    \label{hash_ad}
\end{figure}

We aim to make use of hash encoding to accelerate the convergence of PINNs. 
However, unlike in NeRF applications, the loss function of PINNs requires derivatives of the output field with respect to the input coordinates. 
Since the proposed hash encoding includes a linear interpolation, these derivatives can be discontinuous, which results in inaccurate evaluations, especially near the boundary of the resolution grid, and these potential discontinuous derivatives are more frequent for high resolution levels of the hash encoding.
Taking a simple function $f(x)=sin(x)$ as an example, whose various order derivatives are readily available, we test the performance of automatic differentiation (which is used often in PINNs) on a simple network function of $x$ trained to output the value of $f(x)$. However, this simple network will incorporate a hash encoding layer. 
As shown in Figure~\ref{hash_ad}, we observe that the values of the derivatives based on NN with hash encoding are not accurate and their accuracy highly depends on the hyper-parameters of the hash encoding.
Specifically, we need to choose the coarsest and finest resolution, encoding levels, as well as hash table size, carefully to mitigate the impact of the discontinuities, which will also depend on the collocation points.
Our opinion is that the strength of the hash encoding, which fits well with high-frequency details in high resolution hashing will be negated by this weakness in the calculation of AD. 
For example, using a lot of levels and including the finest of resolution would make the whole neural network (NN with encoding) fit the values at the training sample points, but the resulting function will lack smoothness yielding unstable derivatives. 
This is a direct consequence of the linear interpolation used for the hash vectors.
As a result, the derivative of the NN evaluated via automatic differentiation (AD) using the current implementation of hash encoding is unstable.

In the quest for an efficient implementation, we, instead, use the finite-difference (FD) method to calculate the derivatives, as automatic differentiation is also expensive for higher-order derivatives. Since, the finite difference, owning to its name, calculates derivatives over a finite length, it is relatively immune to point-induced derivative discontinuities. Nevertheless, the accuracy might suffer slightly when dealing with functions with abrupt changes, which is a general weakness of PINNs. 
The FD method is built on the Taylor series expansion. 
Given a grid point $\mathbf{x}_i$, its physical field $u(\mathbf{x}_i)$ can be approximated by limiting the length of its Taylor series expansion, as follows:
\begin{equation} 
    u\left(\mathbf{x}_i + \Delta \mathbf{x}\right)=u\left(\mathbf{x}_i\right)+\left.\Delta\mathbf{x} \frac{\partial u}{\partial \mathbf{x}}\right|_{\mathbf{x}_i}+\left.\frac{\Delta \mathbf{x}^2}{2} \frac{\partial^2 u}{\partial \mathbf{x}^2}\right|_{\mathbf{x}_i}+\cdots.
    \label{equ:taylor}
\end{equation}
Stopping at the second-order accuracy, the finite-difference first- and second-order derivatives are given by:
\begin{equation}
\begin{aligned}
\left.\frac{\partial u}{\partial \mathbf{x}}\right|_{\mathbf{x}_i} & \approx\frac{u\left(\mathbf{x}_i+\Delta \mathbf{x}\right)-u\left(\mathbf{x}_i-\Delta \mathbf{x}\right)}{2 \Delta \mathbf{x}}, \\
\left.\frac{\partial^2 u}{\partial \mathbf{x}^2}\right|_{\mathbf{x}_i} & \approx\frac{u\left(\mathbf{x}_i+\Delta \mathbf{x}\right)-2 u\left(\mathbf{x}_i\right)+u\left(\mathbf{x}_i-\Delta \mathbf{x}\right)}{\Delta \mathbf{x}^2} .
\end{aligned}
    \label{equ:fd-der}
\end{equation}
During the training, the mesh points needed for the derivative calculation should be fed into the NN to get the corresponding field values. 

\begin{figure}[!htb]
  \centering
  \subfigure[T=10]{\label{hash_fd_10}\includegraphics[width=0.32\textwidth]{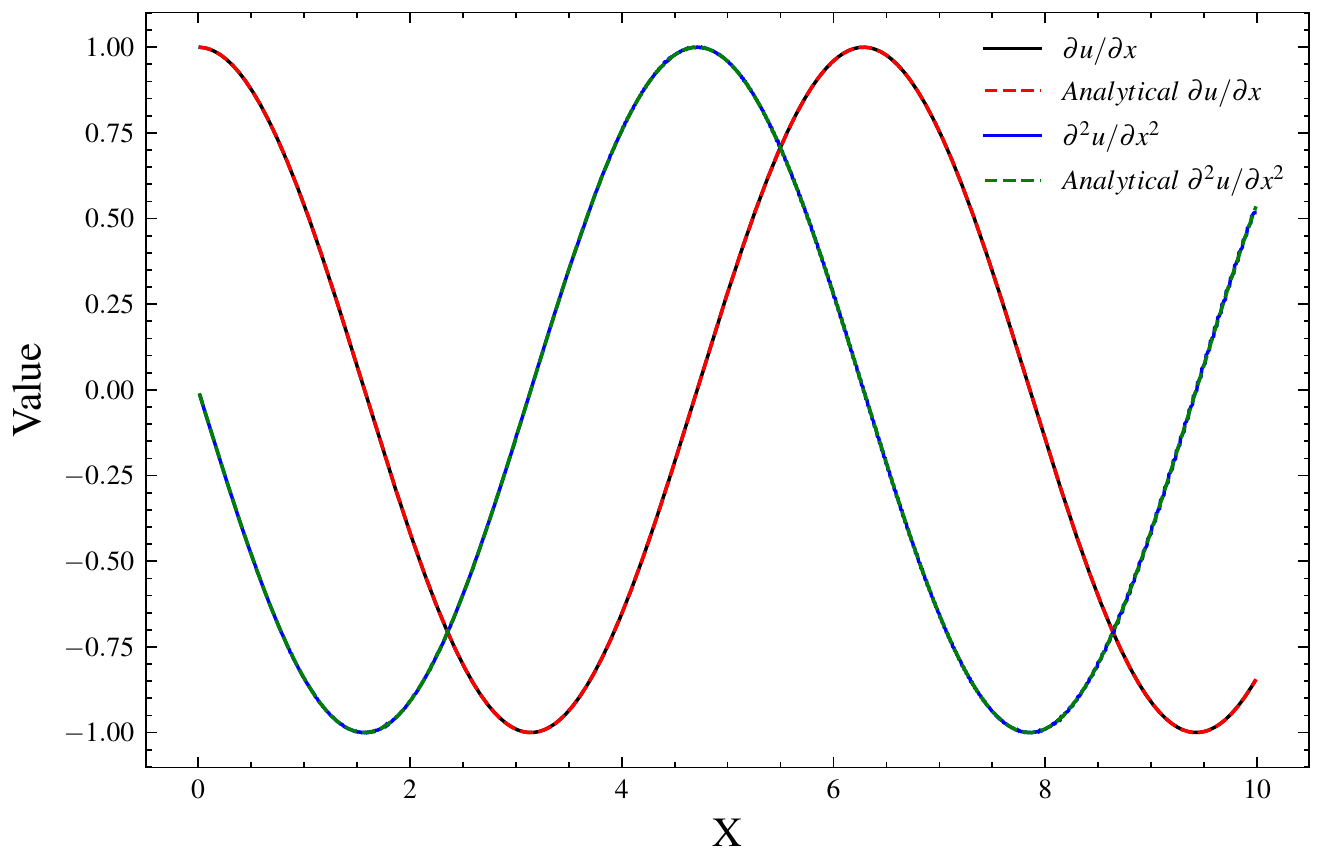}}
    \subfigure[T=8]{\label{hash_fd_8}\includegraphics[width=0.32\textwidth]{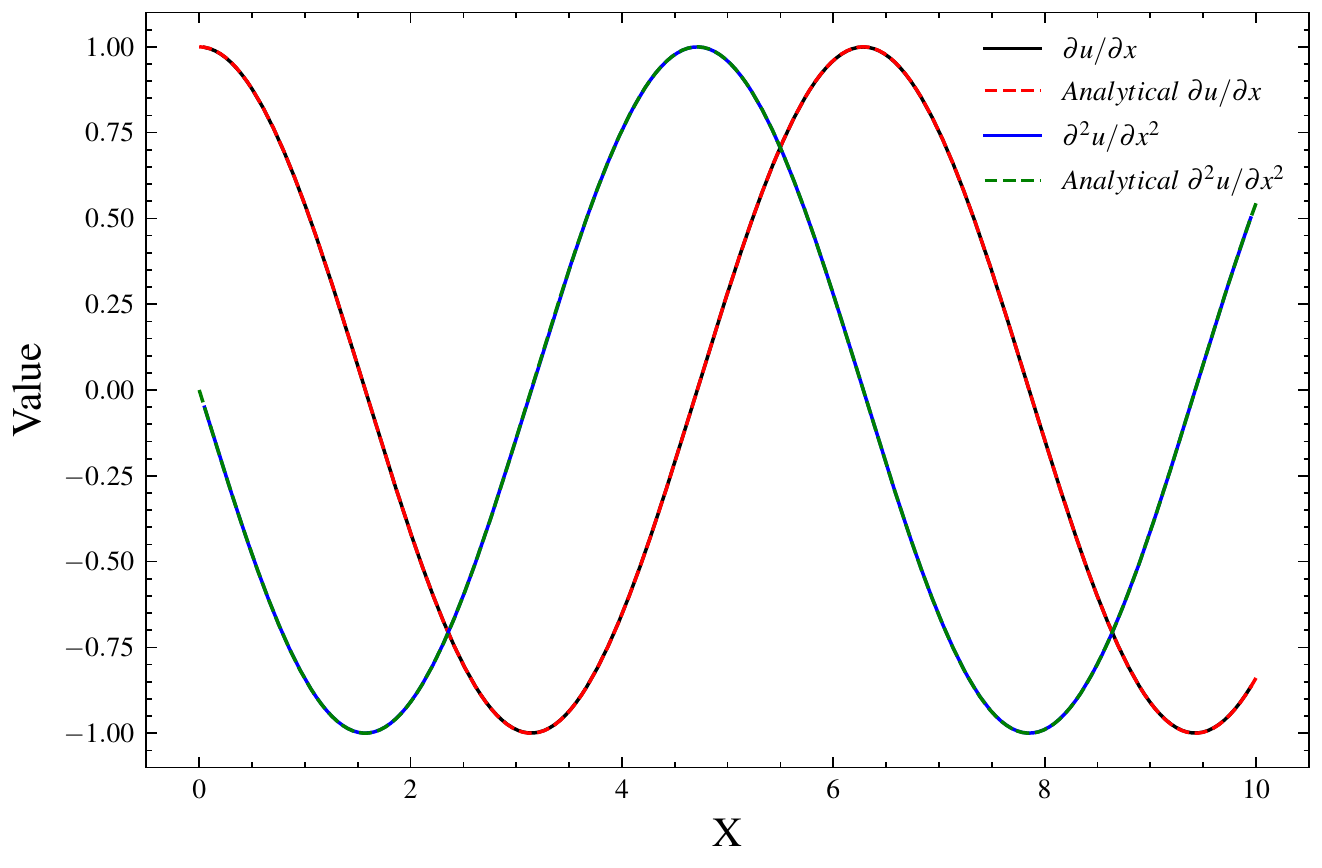}}
    \subfigure[T=4]{\label{hash_fd_4}\includegraphics[width=0.32\textwidth]{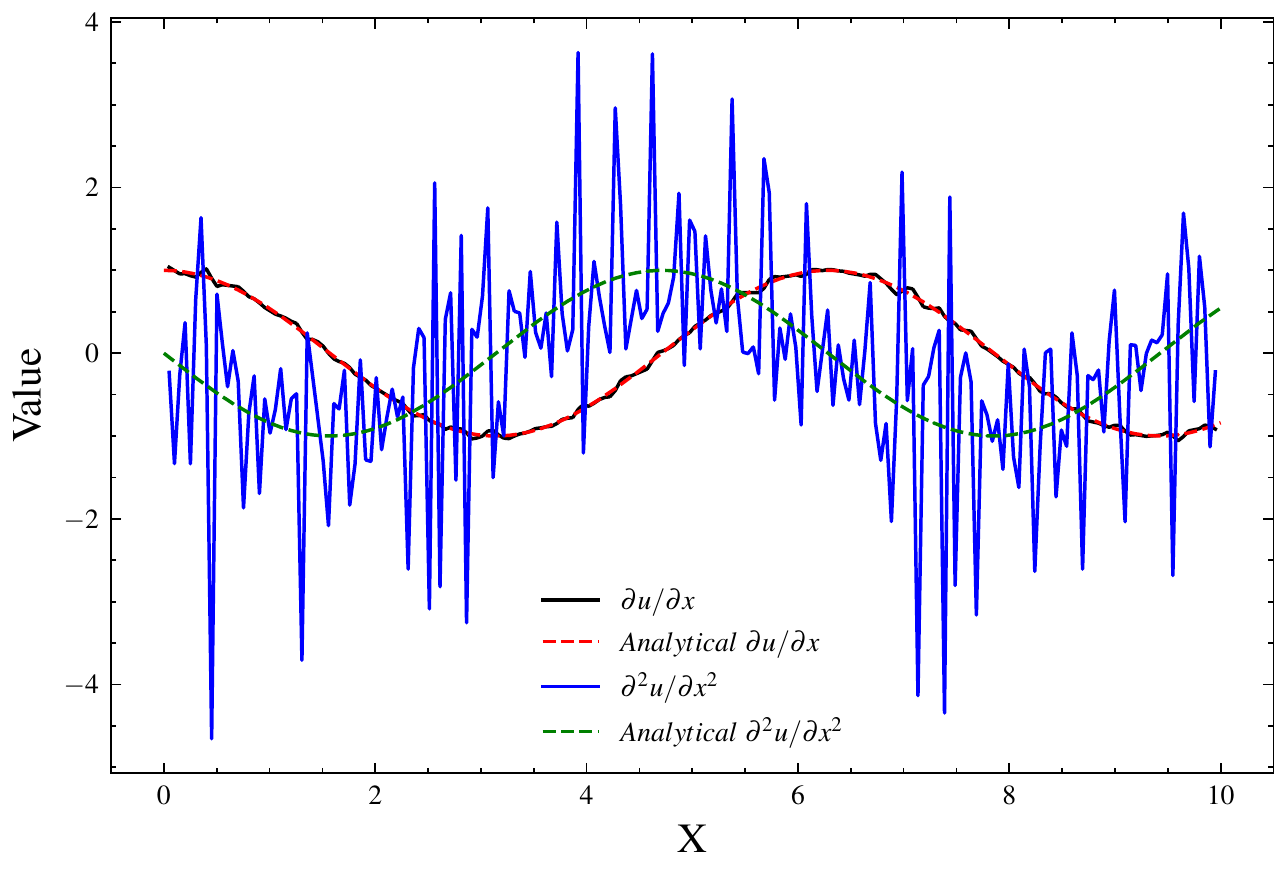}}
    \caption{Illustration of the accuracy of the first- and second-order derivatives calculation by the FD method. We use an NN to fit $f=sin(x)$ with the multi-resolution hash encoding and visualize its first- and second-order derivatives for a hash table size of 10 in (a), and also visualize the derivatives with hash table sizes of 8 and 4 in (b) and (c), respectively.}
\end{figure}

As shown in Figures~\ref{hash_fd_10} and \ref{hash_fd_8}, the derivatives of the NN with hash encoding are generally more accurate than the AD ones, but we still need to carefully pick the encoding hyperparameters.
Here, we show a failure case with FD method in Figure~\ref{hash_fd_4} resulting
from using a small hash table, which forces the NN to learn to distinguish the samples at different locations, yielding a decrease in the accuracy of the derivative calculation (Figure~\ref{hash_fd_4}). 
Compared to the AD method, although the second-order derivative in (b) is not smooth, its trend is consistent with analytical solutions. 
Later we will share our choice of it and other parameters, e.g., the level of resolutions.

\section{Experiments} 
\label{experiments}
In this section, we will showcase the effectiveness of the proposed method through its applications on three well-known partial differential equations (PDEs).
In all cases, we use the MLP architecture as the backbone and Tanh as the activation function. 
For a fair efficiency comparison, we use the FD method to obtain the derivatives for vanilla PINN and PINN with hash encoding. 
We slightly increase the width of the vanilla PINN to make the number of trainable parameters almost equivalent to the PINN with hash encoding.
We train these neural networks with an Adam optimizer and a decaying learning rate in all experiments. 
All experiments used an 80GB NVIDIA A100 GPU card.
Our objective is to demonstrate the gains in efficiency in training PINNs with hash encoding. Thus, for each test, we set a threshold for the solution accuracy (the absolute errors between the predicted solution via the NN and the reference solution) to stop the training and evaluate the approaches based on the number of epochs and the cost of each epoch.
Due to the flexibility of the FD method for derivative calculations, we implemented the NN with tiny-cuda-nn \cite{muller_real-time_2021} framework to accelerate the training even further. 
%More details can be found in Appendix~\ref{app:exp}.

\begin{figure}[!htb]
    \centering    
    \subfigure[]{\label{fig:burger}\includegraphics[width=0.42\textwidth]{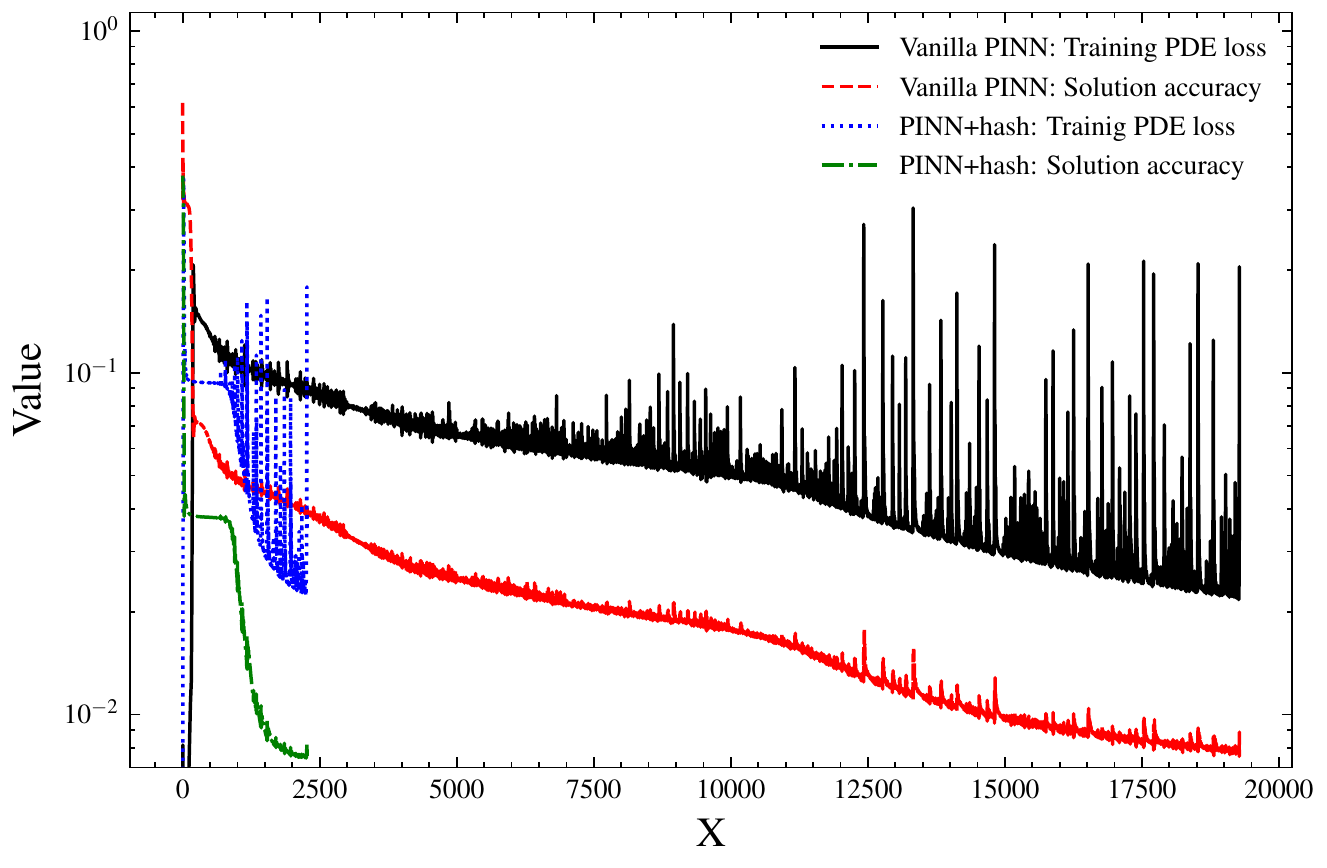}}
    \subfigure[]{\label{fig:burger-vis}
    \includegraphics[width=0.55\columnwidth]{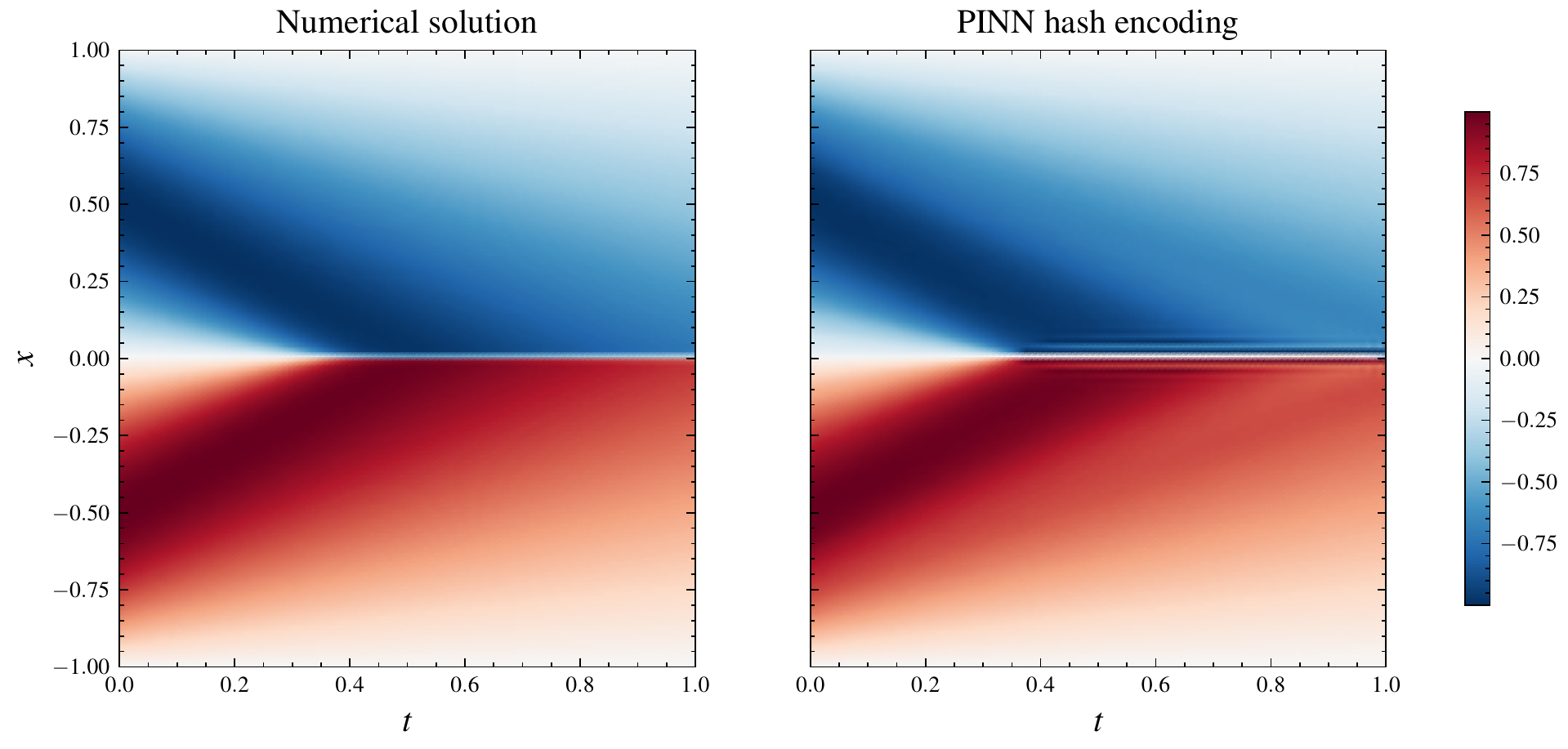}}
    \caption{a) The histories of convergence and testing data errors for the Burgers equation tests, and b) the prediction of PINN with hash encoding and the numerical reference solutions.}
    \label{fig:burger-con-vis}
\end{figure}

\textbf{Burgers equation}. First, we consider a one-dimensional time-dependent equation with a Dirichlet boundary condition representing the one-dimensional flow of a viscous fluid, called the Burgers equation, a widely used benchmark in PINNs. The governing equations are given by \citep{burgers_mathematical_1948}:
\begin{equation}
\begin{aligned}
& \frac{\partial u}{\partial t}+u \frac{\partial u}{\partial x}=\nu \frac{\partial^2 u}{\partial x^2}, \,\,\,\,\, t \in[0,1], \,\,\, x\in[-1,1],  \\
& u(t, -1)=u(t, 1)=0, \\
& u(0, x)=-sin (\pi x),
\end{aligned}
\label{equ:burger}
\end{equation}
where $\nu$ is the viscosity parameter and is set to $\frac{0.01}{\pi}$ here.
For the vanilla PINN, we use an MLP with three hidden layers \{96,96,96\}, while for PINN with hash encoding, we use an MLP of size \{64,64,64\}.
The learning rate is 1e-3 and it is reduced by a factor of 0.8 at the 3000, 5000, 7000, and 10000 epochs.
We consider the numerical solution as a reference to evaluate the accuracy of the predictions.
Figure~\ref{fig:burger} shows the convergence rates of the proposed method and the vanilla PINN using 12800 collocation points.  
If we, as stated earlier, focus on the convergence speed by measuring the number of epochs required to attain a predefined accuracy threshold,  we found that the PINN with hash encoding could reach this target accuracy within less than 2500 epochs, while the vanilla PINN needed almost 20000 epochs. The threshold accuracy considered here admitted a solution that is equivalent to the reference solution (Figure~\ref{fig:burger-vis}).

\begin{figure}[!htb]
    \centering
    \subfigure[]{\label{fig:helmholtz}\includegraphics[width=0.42\textwidth]{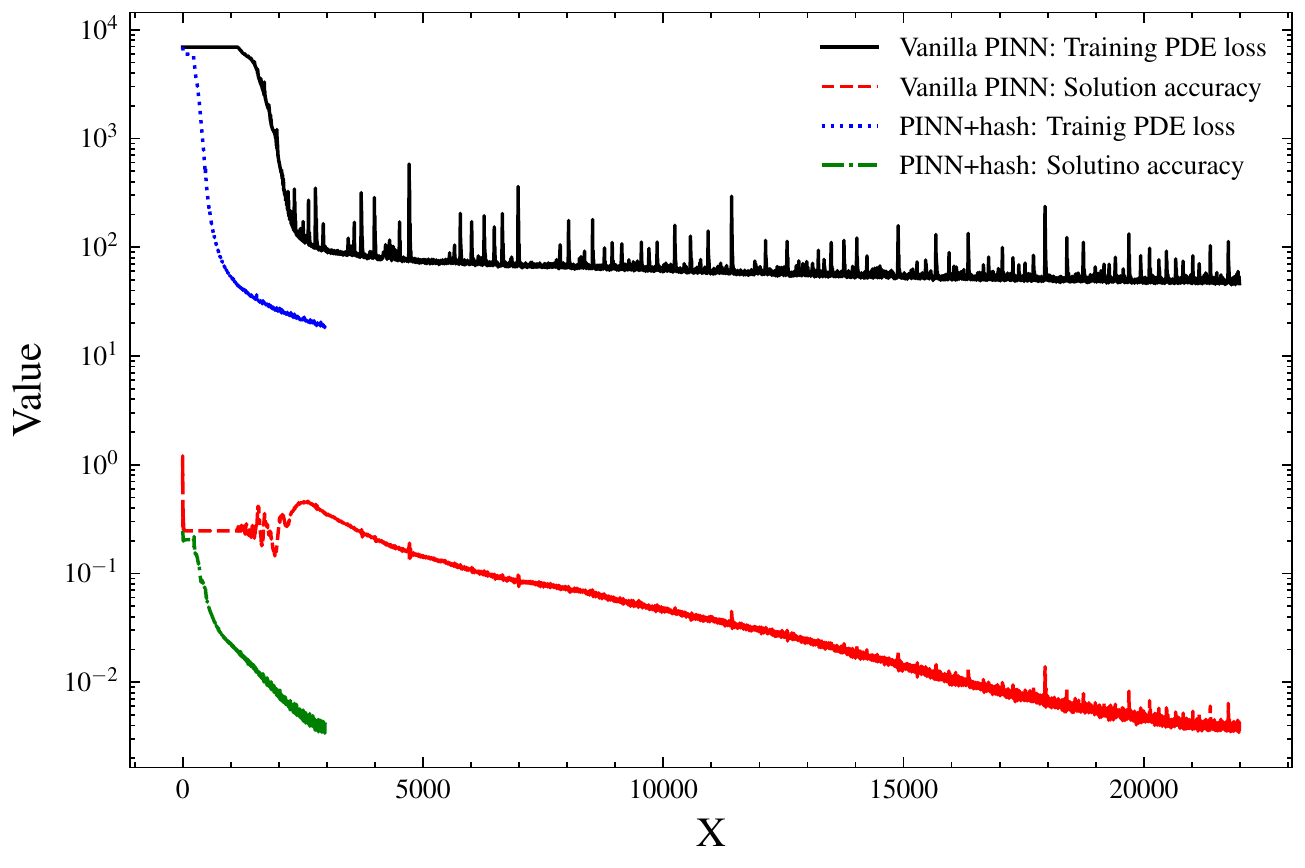}}
    \subfigure[]{\label{helmholtz-vis}\includegraphics[width=0.55\textwidth]{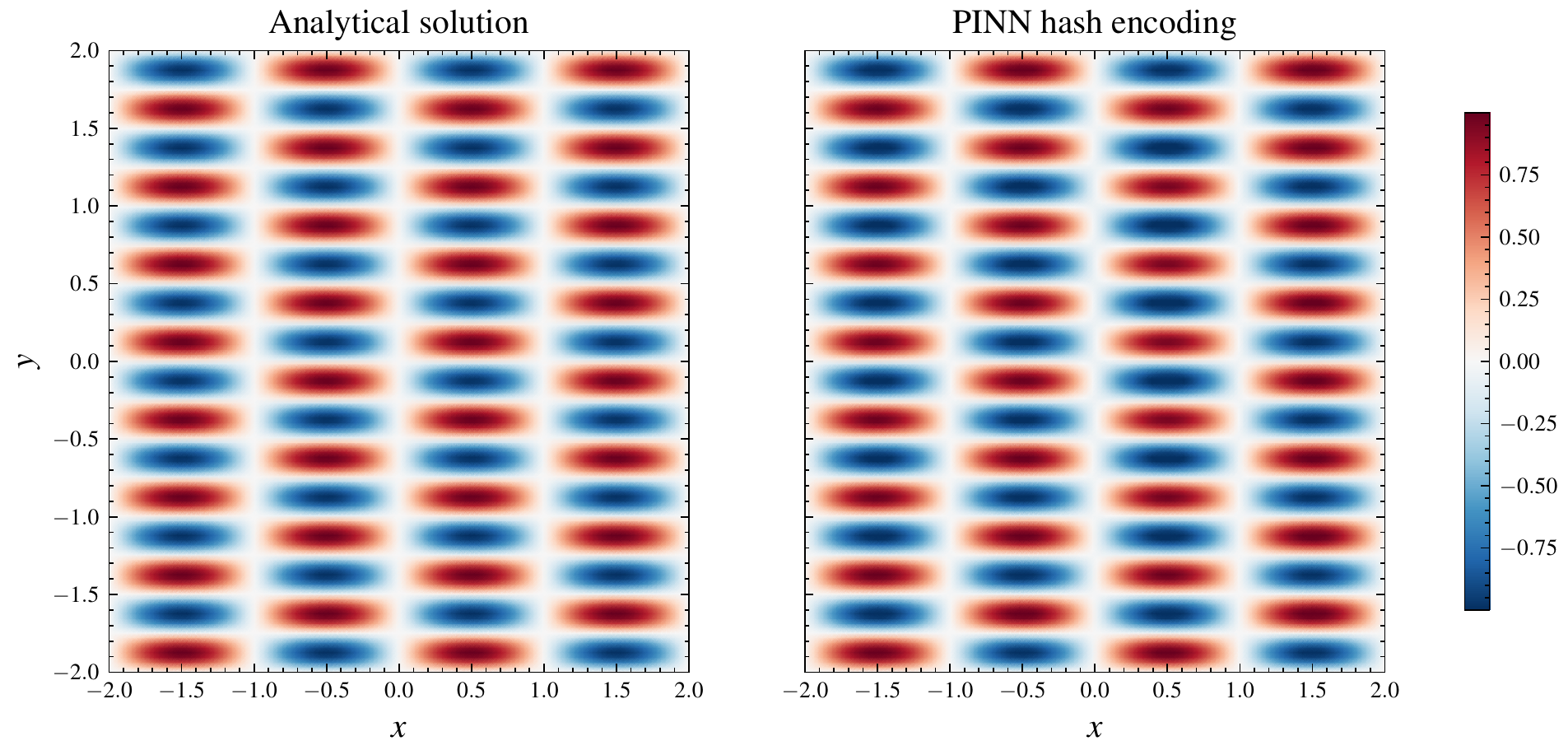}}
    \caption{The histories of convergence and testing data errors for the Helmholtz equation tests, and b) the prediction of PINN with hash encoding and the numerical reference solutions.}
    \label{fig:helmholtz-con-vis}
\end{figure}

\textbf{Helmholtz equation}. Next, we test the PDE on a second-order derivative problem, given by the infamous Helmholtz equation, which describes wave phenomena and has a lot of applications in seismic and electromagnetic fields \citep{riley_mathematical_2002}. Here we consider a simple form of the Helmholtz equation 
\begin{equation}
\begin{aligned}
& \frac{\partial^2 u}{\partial^2 x}+\frac{\partial^2 u}{\partial^2 y}+\lambda u-f(x, y)=0, \\ 
& u(x, 2)=0, \,\,\, u(-2, y)=0, \,\,\, u(x,-2)=0, \,\,\, u(y, 2)=0,\\
& f =-\left(a_1 \pi\right)^2 \sin \left(a_1 \pi x\right) \sin \left(a_2 \pi y\right)\\
&\quad-\left(a_2 \pi\right)^2 \sin \left(a_1 \pi x\right) \sin \left(a_2 \pi y\right)\\
&\quad+\lambda \sin \left(a_1 \pi x\right) \sin \left(a_2 \pi y\right),
    \label{equ:helm
    }
    \end{aligned}
\end{equation}
where $f$ is the source function, $u$ is the wavefield, $\lambda$ is the square of the wavenumber, and $a_1$ and $a_2$ are the parameters to control the sinusoidal nature of the source term.
An analytical solution for this equation exists and is given by \citep{Wang2021under}:
\begin{equation}
u(x,y) = sin(a_1\pi x) sin(a_2\pi y).
    \label{equ:heml-ana}
\end{equation}
In this case, we use an MLP with three hidden layers \{144,144,144\} for the vanilla PINN, while for PINN with hash encoding, we use an MLP of \{128,128,128\}.
The learning rate is 1.5e-3 and it is reduced by a factor of 0.8 at the 3000, 5000, and 7000 epochs. 
We uniformly sample 10000 collection points to train the NN.
The convergence rate for the Helmholtz equation training is shown in Figure~\ref{fig:helmholtz}a. We observe that the PINN with hash encoding admits much faster convergence, and the PDE loss and testing data errors of PINN with hash encoding can still decrease. Nevertheless, the predicted and reference solutions, shown in Figure~\ref{fig:helmholtz}b, look the same.

\begin{figure}[!htb]
    \centering
    \subfigure[]{\label{fig:lid-driven}\includegraphics[width=0.55\textwidth]{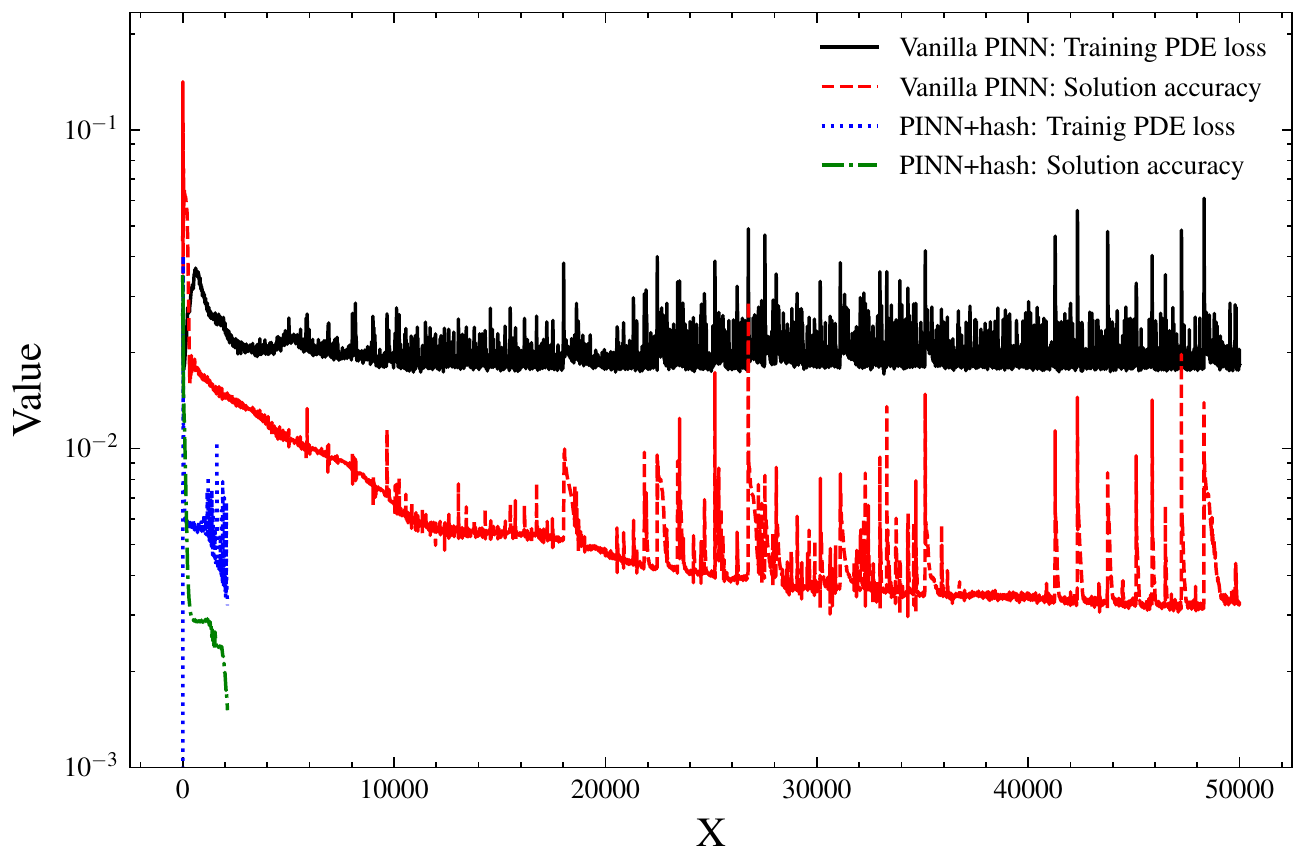}}
    \subfigure[]{\label{lid-driven-vis}\includegraphics[width=0.35\textwidth]{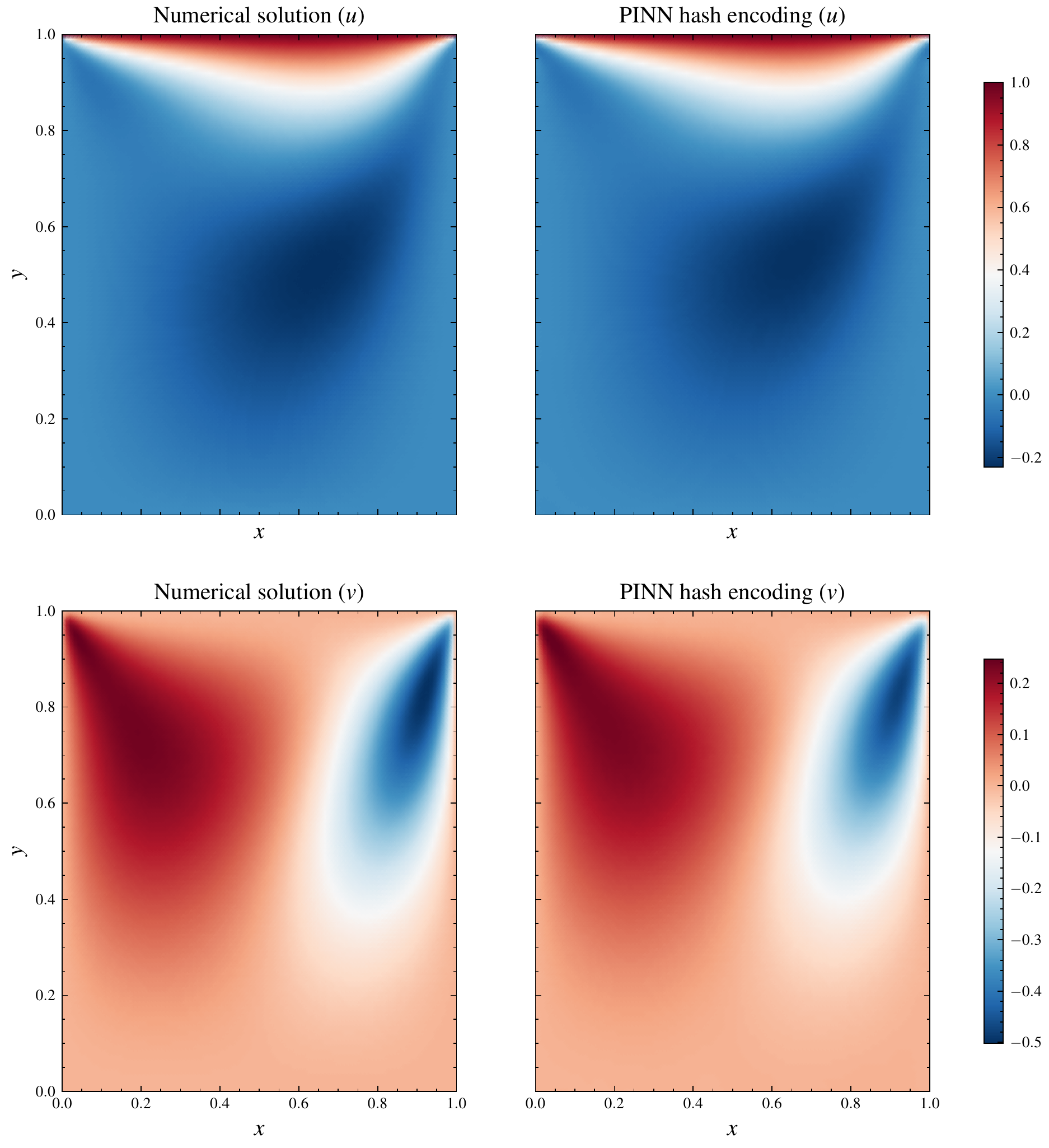}}
    \caption{a) The histories of convergence and testing data errors for the Navier-Stokes equation tests, and b) the prediction of PINN with hash encoding and the numerical reference solutions, where $u$ and $v$ are the horizontal and vertical component of $\vec{u}$.}
    \label{fig:lid-driven-con-vis}
\end{figure}

\textbf{Navier-Stokes equation.} Finally, we test the proposed method on a well-known equation in dynamic fluids, the Navier-Stokes equation. Specifically, we consider the incompressible fluid case, yielding the two governing equations based on mass and momentum conservation, as follows \citep{ethier_exact_1994}:
\begin{equation}
\begin{aligned}
\partial_t \vec{u}(x,y, t)+\vec{u}(x,y,t) \cdot \nabla \vec{u}(x,y,t)+\nabla p & =\frac{1}{Re} \Delta \vec{u}(x,y, t)+f(x,y), & & x \in(0,1)^2, t \in(0, T] \\
\nabla \cdot \vec{u}(x, t) & =0, & & x \in(0,1)^2, t \in[0, T] \\
\vec{u}(x, 0) & =\vec{u}_0(x), & & x \in(0,1)^2
\end{aligned}
\label{equ:ns}
\end{equation}
where $Re$ is the Reynolds number and is set to 100 in our experiments, $\nabla$ is the divergence operator, $\Delta$ is the Laplacian operator, $\vec{u}$ is the velocity field,  $\vec{u}_0$ is the initial velocity field, $p$ is the pressure, and $f$ is the external force, in which we set to zero here.
The vanilla PINN has three hidden layers \{112,112,112\}, in contrast, we use \{64,64,64\} for the PINN with hash encoding.
The learning rate is 1.2e-3 and is reduced by a factor of 0.8 at the 3000, 5000, and 7000 epochs. 
We uniformly sample 10000 collection points to train the NN.
The results are shown in Figure~\ref{fig:lid-driven}, where the reference solution are obtained from numerical solvers. Like previous experiments, the proposed method has fast convergence and can reach the target accuracy (1.5e-3) with only 2270 epochs. However, even with 50000 epochs, the vanilla PINN can not meet the target accuracy. This demonstrates that the proposed method accelerates training and improves accuracy.

\textbf{Training efficiency comparison.} The above experiments demonstrate that PINN with hash encoding can be trained to achieve a good target accuracy within far fewer epochs than the vanilla PINN. In table~\ref{tab:efficiency}, we share a quantitative comparison of the two methods for the three examples we used here. We found that PINN with hash encoding can solve these three famous equations within 30 seconds using a single NVIDIA A100 GPU card.
\begin{table}[!htp]
	\centering
	\caption{The efficiency comparison between the vanilla PINN and PINN with hash encoding.}
	\setlength{\tabcolsep}{0.01\columnwidth}{
	\begin{tabular*}{0.95\columnwidth}{c|c|ccc}
		\toprule[1.5pt]
      Examples & Methods & Time/epoch  & Total cost & Parameter size \\
		\midrule
    \multirow{2}{*}{Burgers equation} & Vanilla PINN & 7.43 ms & 144 s & 21504\\
    & PINN with hash encoding & 7.53 ms & 16.8 s & 20416\\
    \hline
    \multirow{2}{*}{Helmholtz equation} & Vanilla PINN & 7.03 ms & 155 s & 46080 \\
    & PINN with hash encoding & 7.10 ms & 21 s & 42944\\
    \hline
    \multirow{2}{*}{Navier-Stokes equation} & Vanilla PINN & 12.8 ms &640 s & 28672\\
    & PINN with hash encoding & 13.1 ms &27 s & 24608 \\
		\bottomrule[1.5pt]
	\end{tabular*}}
	 \label{tab:efficiency}
\end{table}

\section{Discussion}
\label{discussion}
The neural network solution of a PDE in the form of a function of the coordinates of the solution space allows for continuous representation of the solution and its derivatives rendering opportunities in interpolation, extrapolation, and inversion. The training of such an MLP neural network has proved to be challenging as the high-dimensional topology of the loss function is rather complex, especially for complex solutions. The back-propagation necessary to determine the direction in which we update the neural network parameters encounters a limited imprint of the training samples' coordinates, given by their scalar input values with no neighborhood awareness, in the forward propagation process. Such scalar inputs are also missing any scale resolution information, which is helpful for the network training to have a more locally aware input. This renders the conventional scalar inputs PINNs to have a more point-dependent training. Encoding offers the input a more profound impact on the network, often in the form of a vector representation of the input. With multi-resolution hash encoding, we manage to embed some regional information, beyond the point, and at multi scales, into the inputs. Such area-aware information embedded in the forward propagation improves the topology of the loss function and renders more effective updates to the neural network almost instantly, even if initialized randomly.

The hash encoding implementation we inherited involves learnable parameters given by the lookup feature vectors. These learned parameters are crucial to capturing the multi-resolution nature of the PDE solution with respect to the input coordinates. To optimize the hash encoding, we have to determine the optimal hyperparameters for the lookup feature vector, and that includes the number of resolution levels, the number of feature vectors per level (the size of the hash table), the base resolution, and the size of the feature vector. We set the latter two parameters to 2 and 4, respectively. The other two parameters depend on the expected resolution and complexity of the solution of the PDE. In Burgers equation, we use 9 for both parameters, while in Helmholtz equation, we use 8 for both. Then in the Navier-Stokes equation, we use 9 and 10 for the number of resolution levels and feature vectors, respectively.

The hash function, unlike positional encoding, is not globally differentiable. It includes obvious discontinuities between the hash interval boundaries and the linear interpolation used for the hash vectors. Thus, due to the point nature of the automatic differentiation, this limitation is exaggerated when the hash table is small. as a result, to mitigate this problem, the hash encoding hyperparameters must be chosen carefully. An alternative solution is provided by using the finite-difference scheme to approximate the derivatives of the solution. This approach also admits more efficient calculation of higher order derivatives as compared to AD. Thus, we resorted, in this study, to finite difference calculation of the derivatives. However, we could also utilize higher-order interpolation methods for the Hash vectors as recently proposed by \cite{heo_robust_2023} for NeRf applications, which we will explore in future work.

\section{Conclusion} 
\label{conclusion}
We proposed a physics-informed neural network combined with hash encoding, resulting in a fast convergence to an accurate solution of boundary value problems. Specifically, we investigate the limitations of NN with hash encoding in calculating the derivatives via automatic differentiation and propose using the finite difference method as an alternative to address the issue of the non-smooth gradients, and to help speed up such calculations. We apply our method to a number of examples, including the Burgers equation, Helmholtz equation, and Navier-Stokes equation. With the proposed PINN with hash encoding, the training cost decreases 7 to 24 times. It has the ability to achieve semi-instant training for PINNs, addressing the main drawback of PINN in terms of the training cost. 

\section{Acknowledgement}
The authors thank KAUST for supporting this research and the SWAG group for the collaborative environment. This work utilized the resources of the Supercomputing Laboratory at King Abdullah University of Science and Technology (KAUST) in Thuwal, Saudi Arabia.

\bibliographystyle{plainnat}
\bibliography{references}

%\newpage
%\appendix
%\section{Experiments detail}
%\input{experiment_detail}
%\label{app:exp}

\end{document}